\documentclass[runningheads]{llncs}
\usepackage[T1]{fontenc}
\usepackage{cite}
\usepackage{multirow}
\usepackage{graphicx}
\usepackage{graphicx,verbatim}
\usepackage{amsmath,amssymb,amsfonts}
\titlerunning{PEKA: Parameter efficient knowledge transfer for foundation models}

\begin{document}
\title{Teaching pathology foundation models to accurately predict gene expression with parameter efficient knowledge transfer}
%
\author{Shi Pan\inst{1} \and
Jianan Chen\inst{1} \and
Maria Secrier\inst{1}}
\authorrunning{S. Pan et al.}
%
\institute{University College London, London, UK \\
\email{shi.pan@ucl.ac.uk, jianan.c@ucl.ac.uk, m.secrier@ucl.ac.uk}\\
\url{https://secrierlab.github.io/}}

\maketitle              
\begin{abstract}
Gene expression profiling provides critical insights into cellular heterogeneity, biological processes and disease mechanisms. There has been an increasing interest in computational approaches that can predict gene expression directly from digitalized histopathology images. While image foundation models have shown promise in a variety of pathology downstream analysis, their performances on gene-expression prediction are still limited. 
Explicitly incorporating information from the transcriptomic models can help image models to address domain shift, yet the fine-tuning and alignment of foundation models can be expensive. In the work, we propose Parameter Efficient Knowledge trAnsfer (PEKA), a novel framework that leverages Block-Affine Adaptation and integrates knowledge distillation and structure alignment losses for cross-modal knowledge transfer. We evaluated PEKA for gene expression prediction using multiple spatial transcriptomics datasets (comprising 206,123 image tiles with matched gene expression profiles) that encompassed various types of tissue. PEKA achieved at least 5\% performance improvement over baseline foundation models while also outperforming alternative parameter-efficient fine-tuning strategies. We will release the code, datasets and aligned models after peer-review to facilitate broader adoption and further development for parameter efficient model alignment.


\end{abstract}
\section{Introduction}
Gene expression profiling provides critical insights into cellular heterogeneity, biological processes and disease mechanisms. However, single-cell RNA sequencing (scRNA-seq) \cite{kolodziejczyk2015technology} and spatial transcriptomics (ST) \cite{rao2021exploring} remain costly, time-consuming, and are not routinely implemented in clinical settings \cite{gulati2024profiling}. This has led to an increasing interest in computational approaches that can predict gene expression directly from digitalized histopathology images \cite{wang2025benchmarking}, which are widely used in routine clinical workflows and often considered as the diagnostic gold standard.

Early convolutional neural networks (CNN) and transformers for predicting gene expression were trained on paired bulk RNA-Seq gene expression matrix and whole slide images \cite{schmauch2020deep, jaume2024transcriptomics}. Recently, pathology foundation models \cite{chen2024towards,vorontsov2024foundation,xu2024whole,vorontsov2023virchow}, which employ self-supervised pre-training on large quantities of whole-slide-images (WSI), have been revolutionizing the field and demonstrated superior performance in various downstream tasks including gene expression prediction. 

Despite these advances, the embeddings extracted using image foundation models predominantly capture morphological patterns, which may not optimally align with the underlying biological signals. Apart from pathology findings directly correlating with observable morphological features, for example, mitosis, apoptosis, and cellular differentiation—produce expression \cite{schmauch2020deep}, the vast majority of gene expression changes occur without manifesting visible alterations in cellular morphology. This means histopathology based foundation models may not naturally emphasize the intrinsic dimensions that can link to the transcriptomic domain. It can be considered as an example of domain shift \cite{guan2021domain}, which partly explains the performance bottleneck encountered when attempting to predict gene expression from histopathology foundation model feature spaces. To overcome this limitation, we need to explicitly incorporate information from the transcriptomic domain to guide the adaptation of image features toward dimensions that are more relevant for gene expression prediction.


Single-cell transcriptomic foundation models, such as scFoundation and scGPT \cite{cui2024scgpt, hao2024large}, encode rich biological signals from the transcriptome and have emerged as powerful tools for modeling gene expression data. To address the modality shifting problem and enhance the understanding capabilities of the molecular information, we introduce Parameter Efficient Knowledge trAnsfer (PEKA), a novel framework that leverages Block-Affine Adaptation \cite{kang2024bone} and integrates knowledge distillation and structure alignment losses for cross-modal knowledge transfer. PEKA selectively amplifies dimensions in the image feature space that are most relevant to transcriptomic information, effectively bridging morphological features with their molecular underpinnings. 

We evaluated PEKA in multiple datasets (comprising 206,123 image tiles with matched gene expression profiles) that encompassed various types of tissue, including breast, liver, kidney, and lung cancer subsets from the HEST dataset \cite{jaume2025hest}. Our experiments demonstrate the framework's robustness and generalizability across different histological contexts. Notably, PEKA demonstrates consistent superiority over various baseline models, achieving at least 5\% performance improvement while also outperforming alternative parameter-efficient fine-tuning strategies in gene expression prediction tasks. We will release the code, related datasets, and aligned models after peer-review to facilitate broader adoption and further development for parameter efficient model alignment.

\section{Methods}

\subsection{Problem Formulation}
Given a cell image $\mathbf{X}_{img} \in \mathbb{R}^{H \times W \times C}$ and its corresponding gene expression data $\mathbf{X}_{seq} \in \mathbb{R}^G$, conventional approaches directly employ a pre-trained foundation model $F(\cdot)$ for prediction:
\begin{figure}
    \centering
    \includegraphics[width=\textwidth]{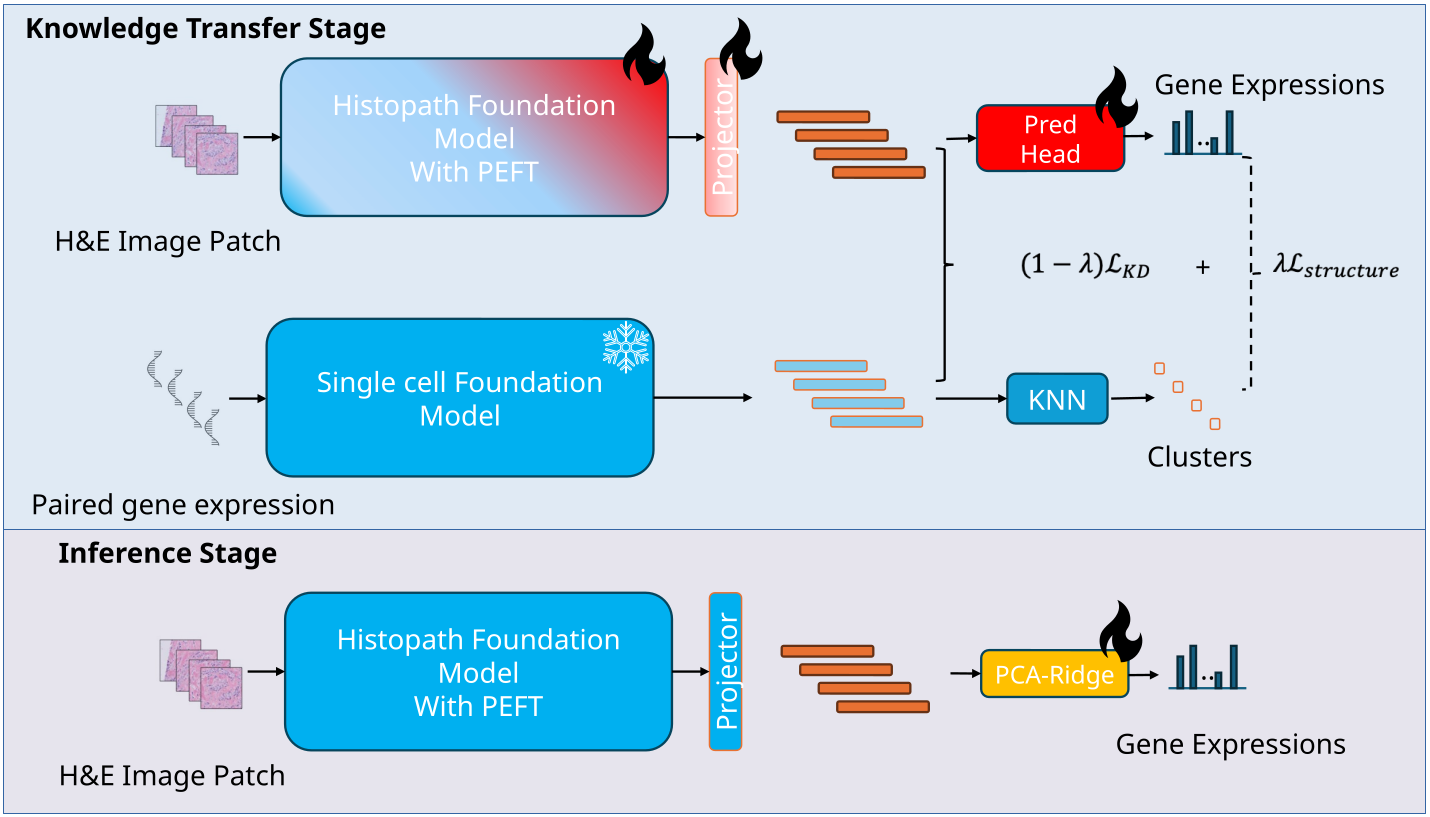}
    \caption{PEKA framework for knowledge transfer between histopathology and transcriptomic foundation models.}
    \label{fig2}
\end{figure}
\begin{equation}
    P(\mathbf{X}_{seq}|\mathbf{X}_{img}) = h(F(\mathbf{X}_{img}))
\end{equation}

where $h(\cdot)$ denotes a task-specific prediction head. However, this direct mapping often struggles to capture complex gene expression patterns because of the significant domain discrepancy between the pre-trained representation space and the gene expression prediction task. We suggest that
this limitation primarily stems from the inherent modality gap between image representations and gene expression patterns, as these models are predominantly optimized for general visual understanding and may lack understanding of molecular features.  

\subsection{Parameter Efficient Knowledge trAnsfer(PEKA)}

To address this challenge, we propose PEKA (Parameter Efficient Knowledge trAnsfer), a plug-and-play framework that enhances the Histopathology Foundation Model's capability in gene expression modeling through efficient knowledge transfer and parameter adaptation (\textbf{Fig. 1}). From an information-theory perspective, our approach bridges the modality gap by maximizing mutual information between the embedding spaces of different models. PEKA achieves this through two key components: knowledge transfer-based modality alignment and parameter-efficient fine-tuning. Specifically, PEKA pipeline consists of two stages: a Knowledge Transfer Stage where H\&E image patches are processed by a histopathology foundation model enhanced with parameter-efficient fine-tuning (PEFT), while matched gene expression data is processed by a single-cell foundation model. The training objective combines direct gene expression prediction ($\mathcal{L}_{KD}$) with structural alignment of feature spaces ($\mathcal{L}_{struct}$), where cluster relationships from transcriptomic embeddings guide the adaptation of image features. In the Inference Stage, only the adapted histopathology model with the PEFT projector is used to process image patches, followed by a 256-dimensional PCA-Ridge regression model to predict gene expression profiles. This approach enables efficient transfer of transcriptomic domain knowledge while modifying around 5\% of the original model parameters.



\subsubsection{Knowledge Transfer with Parameter-Efficient Adaptation}
Given a pre-trained gene expression encoder as the teacher model $T(\cdot)$ and the adapted image encoder as the student model $S(\cdot)$, we formulate the knowledge transfer process as follows:

\begin{equation}
    S(\mathbf{X}_{img}) = F(\mathbf{X}_{img}; \theta_F + \Delta W)
\end{equation}

where $\theta_F$ represents the parameters of the Foundation Model and $\Delta W$ denotes the task-specific parameter updates. The teacher model $T(\cdot)$ provides domain-specific supervision signals to guide the adaptation of the student model. To achieve efficient adaptation while preserving the Foundation Model's general capabilities, we explore various parameter-efficient fine-tuning (PEFT) strategies. The general form of parameter adaptation can be expressed as:
\begin{equation}
\theta_{adapted} = \theta_F + \Delta\mathbf{W} = \theta_F + (\mathbf{W}_{g}\mathbf{B}_{g} + \mathbf{B}_{g} )
\end{equation}
where $\Delta\mathbf{W}$ represents the adaptive weight matrix. To achieve efficient adaptation while preserving structural information, we extend the original low rank adaptation framework by applying block-wise grouping, where $\mathbf{W}_{g}  \in \mathbb{R}^{n \times m}$ is reshaped into blocks of size $b \times b$. These blocks are then strategically grouped to significantly reduce the number of trainable parameters without compromising performance. This grouped block structure yields a closed-form solution equivalent to the Bone formulation \cite{kang2024bone}, which demonstrates superior performance in bridging the modality gap while maintaining parameter efficiency.


\subsection{Optimization target}

The training objective of PEKA comprises three components: knowledge distillation loss, structure alignment loss, and regularization term. The complete loss function can be formulated as:

\begin{equation}
   \mathcal{L}_{total} = \lambda_1\mathcal{L}_{KD} + \lambda_2\mathcal{L}_{struct} 
\end{equation}

where $\lambda_1$ and $\lambda_2$ are weights balancing different losses. Each component serves a specific purpose:

\begin{itemize}
   \item \textbf{Knowledge Distillation Loss} ($\mathcal{L}_{KD}$): Facilitates the transfer of transcriptomic domain knowledge from the teacher model to the student model:
   \begin{equation}
       \mathcal{L}_{KD} = D_{KL}(\text{softmax}(T(\mathbf{X}_{seq})/\tau) \parallel \text{softmax}(F(\mathbf{X}_{img}; \theta_F + \Delta W)/\tau))
   \end{equation}
   where $\tau$ is the temperature parameter controlling the softness of probability distributions.

   \item \textbf{Structure Alignment Loss} ($\mathcal{L}_{struct}$): Preserves the structural relationships in gene expression space:
    \begin{equation}
    \mathcal{L}{struct} = \text{CrossEntropy}(S(\mathbf{X}_{img}), \textit{l})
    \end{equation}
where \textit{l} = $\text{KNN}(\mathbf{X}_{seq})$ generates pseudo-labels by applying k-nearest neighbor clustering in the feature space from teacher model $T(\cdot)$. These structure-aware labels capture the intrinsic neighborhood relationships in the transcriptomic embedding space. By optimizing the cross-entropy loss between the student model's predictions and these pseudo-labels, we encourage the image representations to preserve the structural topology of the teacher model $T(\cdot)$.
\end{itemize}
This training approach ensures effective knowledge transfer while maintaining structural consistency between modalities. In fact, the knowledge distillation process of PEKA results a maximization process of the mutual information between image and transcriptomic domain. 
Single cell large language models (scLLMs) serve as ideal teacher models for PEKA due to their comprehensive understanding of gene expression patterns across diverse cellular contexts. Among available scLLMs, we selected scFoundation \cite{hao2024large} as our teacher model due to its extensive pretraining on over 50 million human single-cell transcriptomic profiles and shows good performance in multiple downstream tasks. 
The embeddings from scFoundation serve as supervision signals during the knowledge distillation process, guiding our image-based student model to learn meaningful representations that align with biologically relevant gene expression patterns. By distilling knowledge from scFoundation, our approach benefits from its extensive pre-training while adapting this knowledge to the spatial context provided by the imaging modality.

\section{Experiments and Results}
\subsection{Dataset Descriptions}
The HEST dataset contains 1,229 whole slide images (WSI) with paired spatial transcriptomic (ST) profiles \cite{jaume2025hest}. To mitigate the effect of different ST technologies and types of disease, and ensure a clean comparison, we selected subsets from the HEST dataset to construct four different benchmarking datasets. The datasets contain Visium ST data of Homo Sapiens with breast cancer (n=30,414 pairs of image tiles and gene expression data), kidney cancer (n=73,813), liver cancer (n=37,168), and lung cancer (n=64,728), respectively. We applied 5-fold cross-validation to compare the performances of our model with other methods.


\subsubsection{Training Settings}
We applied the same quality control steps of scFoundation to filter out low-quality and damaged cells \cite{hao2024large}. Following the designs in HEST, we calculated the top 50 highly-variale-genes (HVG) in each dataset as targets for evluating models, and employed a 256-dimensional PCA followed by a ridge regression classifier to predict gene expressions from foundation model (FM) embeddings. We compared our PEKA model against backbone FMs including Resnet-50 trained on ImageNet \cite{wightman2021resnet}, CtransPath\cite{wang2022transformer}, UNI \cite{chen2024towards} and Hoptimous0 \cite{hoptimus0}. We also compared our method with different strategies for parameter efficient finetuning, including Low-Rank Adaptation (LoRA) \cite{hu2022lora}, AdaLoRA\cite{zhang2023adalora}. 

During the Knowledge Transfer Stage of PEKA, the adapter parameters and MLP weights are updated while keeping the backbone frozen. In the inference stage, we freeze the entire adapted model (backbone with PEFT) and only train a PCA-Ridge regression head for gene expression prediction.

\subsubsection{Evaluation metric}
Pearson correlation coefficient (PCC) is used as the evaluation metric. 
\begin{equation}
r_{xy} = \frac{\sum_{i=1}^{n}(x_i - \bar{x})(y_i - \bar{y})}{\sqrt{\sum_{i=1}^{n}(x_i - \bar{x})^2} \sqrt{\sum_{i=1}^{n}(y_i - \bar{y})^2}}
\end{equation}
where, $x$ and $y$ are the predicted and ground truth gene expression values respectively, $\bar{x}$ and $\bar{y}$ are their respective means, and $n$ is the total number of samples. 

\subsubsection{Hyper parameters} 
We ran knowledge transfer experiments for 50 epochs, using an Adam optimizer with a learning rate of 0.0001. We used $r=256$, $\alpha=32$, and dropout rate of 0.1 for the low rank adaptation of the models. We set $\lambda_1=\lambda_2=0.5$ for the alignment of models. The backbone we selected: H-optimus-0 is a large foundation model, which include 1.1B paramerters. Using PEKA, we achieve parameter-efficient and data-efficient knowledge transfer by only tuning around 5\% of the parameters in the imaging backbone. The alignment takes 12 GPU hours on a single V100 GPU.

\begin{table}[]
\centering
\caption{PEKA significantly outperforms Resnet, CtransPath, UNI and H0 in Pearson correlation (PCC) for gene expression prediction between the imaging and expression embedding spaces.}
\begin{tabular}{lccccc}
\hline
\textbf{Model}       & \textbf{HEST-Breast} & \textbf{HEST-Kidney} & \textbf{HEST-Liver} & \textbf{HEST-Lung}  \\ \hline
\textbf{ResNet-50}   & 0.554           & 0.587           & 0.554          & 0.696          \\
\textbf{CTransPath}  & 0.424           & 0.410           & 0.489          & 0.592          \\
\textbf{UNI}         & 0.617           & 0.673           & 0.576          & 0.708          \\
\textbf{Hoptimous0}  & 0.624           & 0.698           & 0.594          & 0.720          \\
\textbf{PEKA (ours)} & \textbf{0.698}  & \textbf{0.755}  & \textbf{0.654} & \textbf{0.774} \\ \hline
\label{tab1}
\end{tabular}
\end{table}

\subsubsection{Knowledge transfer results}

We trained our models by transfering knowledge from a scFoundation model (teacher) to a Hoptimous0 image foundation model (student) using our parameter efficient knowledge transfer (PEKA) approach. 
\begin{figure}
    \centering
    \includegraphics[width=\textwidth]{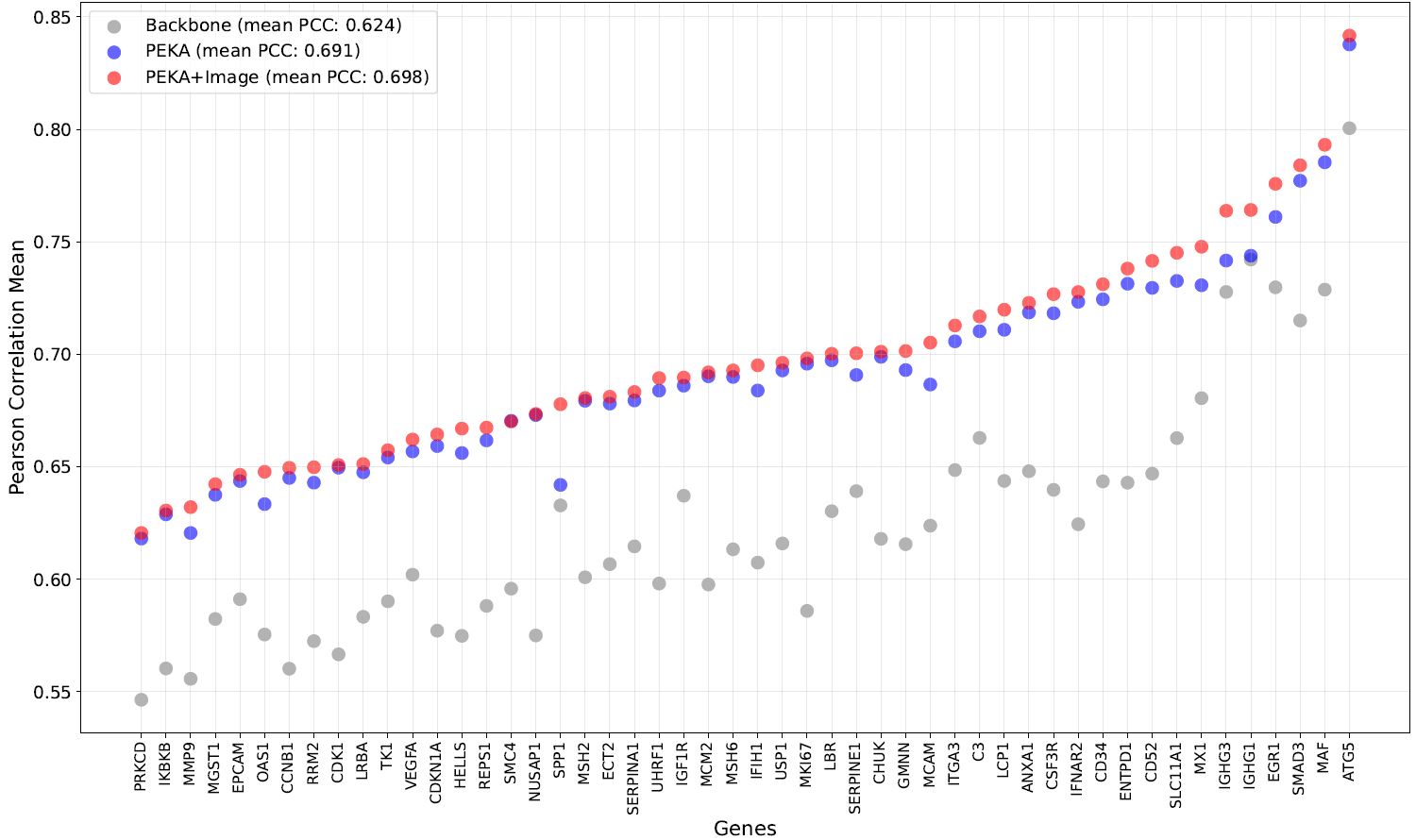}
    \caption{PCC of top 50 highly-variable-genes in the breast cancer dataset.}
    \label{fig2}
\end{figure}

We compared our model with other foundation models in the four datasets derived from HEST. Our method significantly outperforms baseline models, with a 5-7\% absolute increase in PCC compared to the second-best performing model (\textbf{Table. \ref{tab1}}). Surprisingly, CTransPath, which is pretrained on a large cohort of pathology images, has much lower PCC compared to Resnet-50 pretrained on ImageNet. This observation shows the potential risk of foundation models overfitting to one modality, and highlights the importance bridging different modalities with knowledge transfer algorithms. Using the PCC of top 50 highly variable genes in the HEST-Breast dataset as an example, we show that PEKA improves gene expression prediction for all of the genes by a large margin (\textbf{Fig. 2}). Concatenating the embeddings of the aligned model with the embeddings of the pathology image foundation model also consistently improved prediction accuracy, suggesting that the aligned model captured extra information complementary to imaging through the knowledge transfer process.

\begin{table}[]
\centering
\caption{PEKA consistently outperforms other parameter efficient fine-tuning methods correlation coefficient across backbones and datasets. Values in the table are Pearson correlation coefficients between predicted gene expressions and ground truth.}
\begin{tabular}{ccccccccc}
\hline
\textbf{Backbone} & \multicolumn{4}{c}{\textbf{UNI}}                                  & \multicolumn{4}{c}{\textbf{Hoptimus0}}                            \\ \cline{2-9} 
\textbf{Method}   & \textbf{None} & \textbf{LoRA} & \textbf{AdaLoRA} & \textbf{PEKA}  & \textbf{None} & \textbf{LoRA} & \textbf{AdaLoRA} & \textbf{PEKA}  \\ \hline
\textbf{Breast}   & 0.617         & 0.668         & 0.643            & \textbf{0.688} & 0.624         & 0.687         & 0.658            & \textbf{0.698} \\
\textbf{Kidney}   & 0.673         & 0.711         & 0.686            & \textbf{0.729} & 0.698         & 0.734         & 0.715            & \textbf{0.755} \\
\textbf{Liver}    & 0.576         & 0.626         & 0.608            & \textbf{0.654} & 0.594         & 0.601         & 0.586            & \textbf{0.640} \\
\textbf{Lung}     & 0.708         & 0.752         & 0.734            & \textbf{0.774} & 0.720         & 0.738         & 0.716            & \textbf{0.774} \\ \hline
\label{tab2}
\end{tabular}
\end{table}

\subsubsection{Comparison with other PEFT methods}

We compared our method with three other settings of parameter efficient fine-tuning (PEFT) including \textbf{None} (No fine-tuning), \textbf{LoRA} (Low Rank Adaptation) \cite{hu2022lora} and \textbf{AdaLoRA} \cite{zhang2023adalora}. In all four datasets, we found that all PEFT strategies significantly improves PCC of gene expressions compared to \textbf{None}. Our approach always provides the largest performance gain, with \textbf{LoRA} being the second.

\section{Conclusions}
In conclusion, we propose PEKA, a parameter efficient knowledge transfer workflow for aligning foundation models trained on different modalities. Foundation models fine-tuned using PEKA achieved state-of-the-art performance in predicting gene expressions from pathology image tiles across four different tissue types. From our results, PEKA can effectively bridge the modality gap between histopathology images and gene expression data, enabling more accurate molecular predictions from morphological features. We believe PEKA and the aligned model will serve as powerful tools for researchers and clinicians to extract valuable transcriptomic insights from routinely available histopathology images. Furthermore, PEKA provides a strong baseline for future advancements in cross-modal knowledge transfer, particularly in the medical domain, where paired multi-modal data are scarce. In the future, we plan to leverage the data-efficiency and parameter-efficiency of PEKA to align other modalities  such as proteomics\cite{miao2021multi}, metabolomics \cite{johnson2016metabolomics}, or epigenomics \cite{wang2018epigenomics} across cancer types and evaluate its utility in precision oncology applications, for example cancer subtyping and survival analysis.

%
%
%
\newpage
\bibliographystyle{splncs04}
\bibliography{peka.bib}
\end{document}